\documentclass[letterpaper, 10 pt, journal, twoside]{IEEEtran}

\usepackage{graphicx}
\usepackage{amsmath} 
\usepackage{amssymb}  
\usepackage{amsthm}  
\usepackage{color}
\usepackage[linesnumbered,lined,ruled]{algorithm2e}
\usepackage{subcaption}

\newcommand{\RAL}[1]{\textcolor{black}{#1}}
\newcommand{\RALF}[1]{\textcolor{black}{#1}}
\newtheorem{lem}{Lemma}

\begin{document}

\title{
CARPAL: Confidence-Aware Intent Recognition for Parallel Autonomy
}

\markboth{IEEE Robotics and Automation Letters 2021. Author Version.}
{Huang \MakeLowercase{\textit{et al.}}: CARPAL: Confidence-Aware Intent Recognition for Parallel Autonomy}


\author{Xin Huang$^{1,2}$, Stephen G. McGill$^{1}$, Jonathan A. DeCastro$^{1}$,\\ Luke Fletcher$^{1}$, John J. Leonard$^{1,2}$, Brian C. Williams$^{2}$, Guy Rosman$^{1}$
\thanks{This work was
supported by Toyota Research Institute (TRI). This article solely reflects the opinions and conclusions of its authors and not TRI or any other Toyota entity. } 
\thanks{$^{1}$Toyota Research Institute, Cambridge, MA 02139, USA 
}%
\thanks{$^{2}$Computer Science and Artificial Intelligence Laboratory, Massachusetts Institute of Technology, Cambridge, MA 01239, USA
        {\tt\footnotesize xhuang@csail.mit.edu }}%
}

\maketitle

\begin{abstract}
Predicting driver intentions is a difficult and crucial task for advanced driver assistance systems. 
Traditional confidence measures on predictions often ignore the way predicted trajectories affect downstream decisions for safe driving.
In this paper, we propose a novel multi-task intent recognition neural network that predicts not only probabilistic driver trajectories, but also utility statistics associated with the predictions for a given downstream task.
We establish a decision criterion for parallel autonomy that takes into account the role of driver trajectory prediction in real-time decision making by reasoning about estimated task-specific utility statistics.
We further improve the robustness of our system by considering uncertainties in downstream planning tasks that may lead to unsafe decisions.
We test our online system on a realistic urban driving dataset, and demonstrate its advantage in terms of recall and fall-out metrics compared to baseline methods, and demonstrate its effectiveness in intervention and warning use cases.

\begin{IEEEkeywords}
Intelligent Transportation Systems, Computer Vision for Transportation, Safety in HRI
\end{IEEEkeywords}

\end{abstract}

\section{Introduction}
\label{sec:intro}
\IEEEPARstart{I}{n} recent years, advanced driver assistance systems (ADAS) have played an important role in improving driving safety. One important aspect of these systems is the prediction of future driver intentions. Many motion prediction approaches have been proposed \cite{lee2017desire,cui2018multimodal,huang2019uncertainty,huang2019diversity,chai2019multipath} to generate accurate probabilistic motion predictions for vehicles. Leveraging the prediction uncertainties in decision making towards safe driving systems with a low error rate still remains an open challenge, due to the data-driven nature of these approaches.

\begin{figure}
    \centering
    \includegraphics[width=1.0\columnwidth]{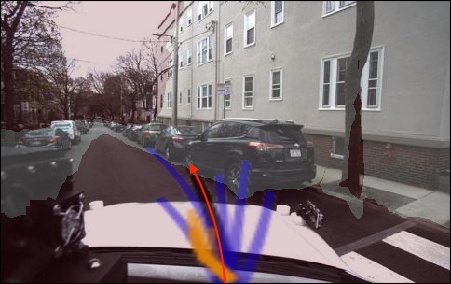}
    \caption{A motivating example, where the observed future driver trajectory (\RALF{in red}) leads to a near collision with a parked car. While a task-agnostic predictor fails to intervene due to the high uncertainties in its predictions (in blue), CARPAL intervenes with a backup planner based on estimated task-specific utility uncertainties. The planner generates safe backup trajectories (in orange) based on a safety utility map (shaded area) and a driver intention utility map.
    }
    \label{fig:motivating_example}
\end{figure}

Solving this challenge could benefit from better insight on how prediction affects downstream decision making systems. For example, \cite{huang2019uncertainty} proposes a driver trajectory predictor that reasons about predictor errors in parallel to generating predictions, which can be used to decide whether the predictions are trustworthy. The predictor errors, however, are \emph{task-agnostic}, in that they do not relate to the prediction effect on downstream tasks, such as planning and collision avoidance. 
In this paper we extend task-agnostic approaches to \emph{task-specific} confidence measures, and demonstrate this idea within parallel autonomy \cite{schwarting2017parallel}, an extension of shared control \cite{abbink2012haptic,saleh2013shared}, that leverages a perception stack and a prediction stack to monitor the safety of a human-driven vehicle, and a planning stack to intervene if necessary to avoid near collisions. Our proposed approach, \emph{Confidence-Aware intent Recognition for ParALlel autonomy (CARPAL)}, not only predicts future driver trajectories, but also estimates trajectory utilities and their task-specific uncertainties through a deep neural network regressor to decide whether it is necessary to take over human control. The regressor enables \emph{real-time} online decision making, by avoiding prohibitively expensive online computation of utilities. 

We define the trajectory utility based on two widely accepted principles in parallel autonomy \cite{schwarting2017parallel,shia2014semiautonomous}: safety and similarity to driver intention.
In our work, we chose keeping a minimum safe distance from obstacles as a safety criteria, though other choices are possible. 
Intention similarity is also crucial for a parallel autonomy system, as we want to follow driver intentions as close as possible, subject to safety. To measure intention similarity, we look at how close are the trajectories to what the human driver could have done, in terms of the predicted distribution.

When the predicted trajectories seem risky, a parallel autonomy system should generate a backup plan that follows driver intentions. 
The backup planner generates safe and physically feasible motion plans according to driver intentions. Different from existing shared control literature \cite{schwarting2017parallel,reddy2018shared,anderson2012constraint,alonso2014shared,erlien2015shared}, our work incorporates not only uncertainty estimates relevant to utility of the driver predictions, but also the uncertainties from the downstream parallel autonomy tasks, such as backup planning, to improve decision making. This allows us to improve safety and reduce false alarms when compared to task-agnostic confidence estimates.

A motivating example is illustrated in Figure~\ref{fig:motivating_example}, where our probabilistic motion predictor generates a set of prediction samples over the next three seconds in blue, and the observed future driver trajectory (visualized in red) indicates a near collision to a parked car on the right.
Due to weak coupling between the prediction uncertainty and the scene information, a task-agnostic predictor such as \cite{huang2019uncertainty} may not intervene based on its confidence level in terms of prediction accuracy.
On the other hand, our proposed online method \emph{CARPAL} provides better task-specific confidence estimates of its predicted trajectories, including the trajectory utilities and their uncertainties measured in this specific scene. Such estimates result in a better decision by taking over driver control using a safe backup plan (orange trajectory), generated by an auxiliary backup planner based on a safety utility map (shaded area) and a driver intention utility map.




\textbf{Contributions} 
i) We propose an online \emph{confidence-aware predictor} that predicts probabilistic future driver trajectories, in addition to regressing \textit{task-specific} utility statistics from its predictions with respect to downstream tasks.
The utility regressor is embedded within the predictor and provides utility estimates to support \textit{real-time decision making}.
ii) We define a utility function in parallel autonomy and a backup planner conforming to this utility function, and propose a binary intervention decision function that leverages utility estimates from predictions and the downstream stack in parallel autonomy to balance safety and following driver intentions. We provide an upper bound of the utility uncertainty of parallel autonomy in terms of the regressed utility estimates.
iii) We demonstrate the effectiveness of our system in augmented realistic risky scenarios that would require interventions or warnings by ADAS. We show how our approach achieves better decision-making results, in terms of recall and fall-out, compared to baseline approaches in the context of parallel autonomy.


\section{Related Work}
\label{sec:related_works}
Parallel autonomy is a vehicle shared-control framework \cite{schwarting2017parallel,saleh2013shared} that monitors driver actions and intervenes before an unsafe event could happen. Many existing methods on parallel autonomy assume a deterministic future driver trajectory or simple dynamics given current driver commands \cite{schwarting2017parallel,reddy2018shared,anderson2012constraint,alonso2014shared,erlien2015shared}, which is insufficient in safety critical driving scenarios, such as turning and merging. In \cite{shia2014semiautonomous}, although multiple trajectories are predicted, only one trajectory is used to evaluate driver risk. Besides, these approaches assume that the backup intervention system has access to perfect environment information and thus always produces safe plans. However, due to imperfect perception, there also exist uncertainties in the downstream stack in parallel autonomy, such as the backup planner. In this work, we propose a robust parallel autonomy system that considers trajectory uncertainties due to both driver actions and perception/planning limitations.



Our approach interacts with many existing works that provide probabilistic models of road agent trajectories, e.g. \cite{deo2018convolutional,wiest2012probabilistic,IvanovicPavone2019,kim2017probabilistic}.
It further relates to uncertainty modeling in learning-based methods; it is important to estimate the confidence of the predictors themselves. Along this line, \cite{huang2019uncertainty} utilizes in-vehicle data from a front camera and CAN bus to produce probabilistic driver trajectory predictions and estimate the prediction error. Beyond prediction, \cite{richter2017safe,amini2018variational} estimate the confidence level in imitation learning via error detection on reconstructed input, and \cite{wulfe2018real} regresses the risk associated with driving scenarios. We extend \cite{huang2019uncertainty} and propose a multi-task model to estimate task-specific utility scores associated with predictions to facilitate real-time decision making in parallel autonomy. The transition from task-agnostic uncertainty in \cite{huang2019uncertainty} to task-specific measures is motivated by ideas in value of information \cite{howard1966information}, and recent approaches for task-specific uncertainty approximants \cite{rosman2018task}. 

A challenge in parallel autonomy is to determine whether to intervene on behalf of the driver based on the confidence of driver risk evaluation. A low confidence level indicates that the predictor may not be able to provide accurate predictions, which may lead to a false alarm if the driver is driving safely but gets intervened. 
In \cite{fisac2018probabilistically,fridovich2019confidence}, the authors propose a prediction-aware planning system that infers the confidence of predictions by maintaining a Bayesian belief over model variances, which is updated through tracking agent actions against a set of known goal states. 
In our work, we measure the confidence level of predictions by regressing the variances of utilities over sampled predicted trajectories directly through a neural network.
In \cite{lee2018safe}, the authors measure the uncertainties of an end-to-end control task by computing the variances of sampled control outputs, and decide whether to give control back to a backup planner by comparing against a dynamically optimized threshold, as opposed to a fixed task-agnostic threshold. Using a similar idea in parallel autonomy, we compare the driver utilities with the utilities of alternative driving options, such as a backup planner, to facilitate task-specific decision making. While there are many planner options, we demonstrate our framework through an example backup planner that finds optimal paths by avoiding obstacles identified by a perception system with sensor noises, and follow driver intentions subject to safety.

\begin{figure*}[t]
    \centering
    \includegraphics[width=1.0\linewidth]{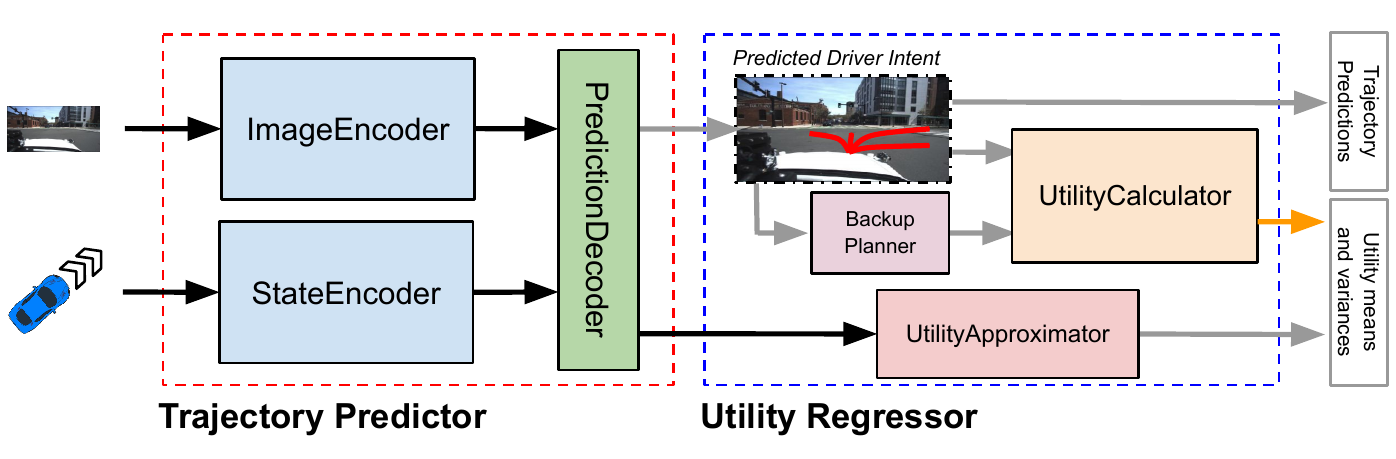}
    \caption{Architecture diagram of \emph{CARPAL}. The trajectory predictor generates probabilistic vehicle trajectories by taking camera images and vehicle states as input. A utility calculator provides at training time supervisory utility estimates for both predicted samples and planned trajectories from a backup planner. These utility cues are used as target values for a utility approximator, which estimates means and variances of prediction and planner utilities to enable real-time decision making at test time. \RALF{Black arrows indicate DNN computation; Grey arrows indicate additional output generation; Orange arrow is for training time only.}}
    \label{fig:architecture_diagram}
\end{figure*}

\begin{figure}[t]
    \centering
    \includegraphics[width=1.0\linewidth]{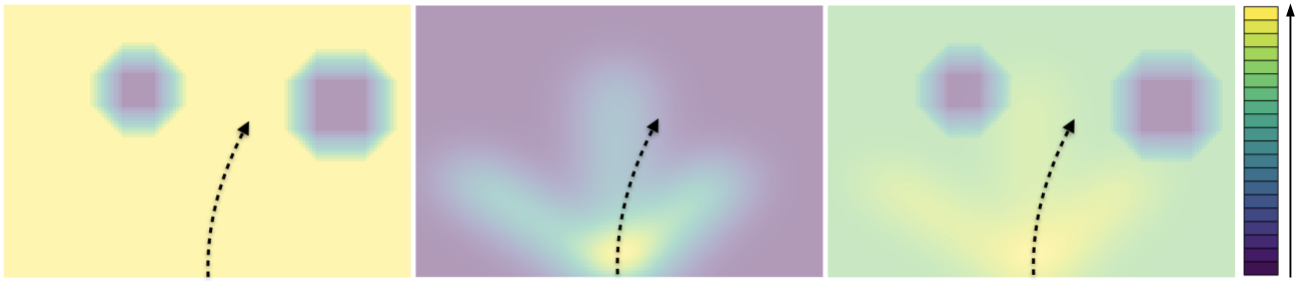}
    \caption{\RAL{Simplified utility map examples used to compute utilities with a given (predicted or planned) trajectory illustrated as the dotted arrow.} Left to right: safety, intent, and overall utility. \RAL{In each map, utility values range from low (purple) to high (yellow). Color scales are different between the maps.}}
    \label{fig:utility}
\end{figure}

\section{Problem Formulation}
\label{sec:problem_formulation}

Given the observed data from the driver vehicle, including past trajectories $\{(x_t,y_t)\}_{t=-T_p+1}^0$ over a fixed horizon $T_p$, CAN bus inputs (steering angle, acceleration) from the vehicle control module, velocity states, and front camera images, we wish to predict future driver trajectories and their utilities, as well as utilities of alternative options to facilitate decision making in parallel autonomy. The trajectories are represented by a probabilistic distribution $P(\mathcal{T} | X)$, where $\mathcal{T}$ denotes future trajectories $\{(x_t,y_t)\}_{t=1}^{T_f}$ up to a fixed horizon $T_f$ and $X$ denotes observed data. 

Several definitions are available for utility in parallel autonomy. The utility $u$ in this work is defined based on the two widely used principles in parallel autonomy \cite{schwarting2017parallel,shia2014semiautonomous}, that capture safety and alignment with driver intentions, as introduced in Section~\ref{sec:intro}. \RALF{The utility $r(s)$ for a given position $s$ is composed of a safety utility term and a driver intention utility term:}
\RALF{
\begin{eqnarray}
    r(s) &=& r_{O}(s) + \alpha r_I(s), \label{eq:utility_all} \\
    r_{O}(s) &=& \text{sigmoid}(\min_{o \in \mathcal{O}} ||s - o||^2), \label{eq:utility_obs} \\
    r_{I}(s) &=& \log P(s|I),
\end{eqnarray}}
\RALF{where the safety utility term $r_{O}$ represents the minimum Euclidean distance to a set of obstacles $\mathcal{O}$, after being normalized by a sigmoid function; the driver intention utility term $r_I$ represents the likelihood of matching human intentions $I$; $\alpha$ is the weighting coefficient to balance the two utilities. The safety utility is normalized by a sigmoid function, because the distance to obstacles matters more when it is small (e.g., in the linear region of the sigmoid function), which indicates a near collision. The human intentions, approximated by a density map generated from sampled predicted human trajectories from a pre-trained network, indicate a set of spatial locations the driver intends to traverse.} Figure~\ref{fig:utility} illustrates examples of utility maps, including a \emph{safety utility map} with two obstacles, a \emph{driver intention utility map} with three potential intended trajectories, and an overall utility map.

\RALF{We can compute the utility of a trajectory $\tau$ including a sequence of $T$ positions by taking the mean over the utilities of each position:}
\begin{equation}
    \RALF{u(\tau) = \frac{1}{T} \sum_{t=1}^T r(\tau^{(t)}).} \label{eq:utility_traj}
\end{equation}

Given the trajectory prediction distribution, we sample a set of future human trajectory prediction samples $\mathcal{T}_H$, and define the mean and variance of their utilities as follows:
\begin{equation}
   \mu_H = \mathbb{E}_{\mathcal{T}_H} [u(\tau_H)], \quad
    \sigma^2_H = \text{var}_{\mathcal{T}_H} [u(\tau_H)].
\end{equation}
The mean value indicates the expected utility of the predicted samples, and the variance measures the uncertainties associated with the utilities.
Similarly, we define the mean and variance from a set of backup planner trajectories $\mathcal{T}_P$ that are designed to guarantee driver safety in risky situations given upstream perception information:
\begin{equation}
    \mu_P = \mathbb{E}_{\mathcal{T}_P} [u(\tau_P)], \quad
    \sigma^2_P = \text{var}_{\mathcal{T}_P} [u(\tau_P)].
\end{equation}

In a binary decision setting for parallel autonomy, the intervention decision function $D$ depends on the utilities of predicted driver behaviors and planned backup trajectories:
\begin{equation}
    D(\mathcal{T}_H, \mathcal{T}_P) = F(\mu_H, \mu_P, \sigma^2_H, \sigma^2_P).
\end{equation}
The decision function returns either a positive decision, which is to take over control using the autonomy trajectory from a backup planner, or a negative decision, which is to keep the human driver operating the vehicle. 



\section{Method}
\label{sec:method}
In this section, we describe our confidence-aware prediction system CARPAL depicted in Figure~\ref{fig:architecture_diagram}, where we first introduce a trajectory predictor based on deep neural networks that outputs distribution over future vehicle positions given input data, and then show how we train a utility approximator model to regress the utility statistics by computing the target supervisory values from training data. Finally, we present how the regressed values are used in a task-specific parallel autonomy system to improve driver safety.
\subsection{Variational Trajectory Predictor}
\label{subsec:predictor}
The trajectory predictor consists of two parts. In the first part, we encode inputs from different sensors using a series of deep neural networks. For instance, the camera image is processed with a pre-trained VGG network \cite{simonyan2014very} with the last classification layer removed to extract road features and drivable areas. \RALF{We pick VGG for simplicity and reproducibility -- while it has difficulties to handle challenging tasks such as detecting fences, we defer a more robust perception system for future work.} The vehicle states, including steering wheel angle, acceleration, past trajectory, and linear and angular velocities, are fed into individual fully connected child networks to produce encoded states. The encoded states from each child encoder network are concatenated together into an embedding layer, which is fed into a trajectory decoder that consists of a series of fully connected networks to generate Gaussian parameters over trajectory coefficients. Similar to \cite{huang2019uncertainty}, we assume that the trajectories can be projected onto a second-order polynomial basis, and represent them as a set of projection coefficients. More details on the network architecture can be found in \cite{huang2019uncertainty}.

\subsection{Utility Regressor}
In CARPAL, we utilize a utility approximator to estimate utility values of predicted trajectories and planned backup trajectories based on a deep neural network to enable online decision making. In order to obtain the supervisory utility cues to train the network, we calculate ground truth acausal utility values that are only available at training time given results from an upstream perception stack, and build a utility-based planner to provide backup safe trajectories given the perception information and driver intention estimates. 

\subsubsection{Ground Truth Utility Calculation}
\label{sec:utility_computation}
We start by calculating the utility values defined in Eq.~\eqref{eq:utility_traj} given ground truth information from data, which will be used as supervisory cues to train a regression deep neural network. 

The obstacles in the \emph{safety utility map} can be obtained from an upstream perception stack. For simplicity and ease of reproducibility, we model the perception stack by a system that identifies obstacles from the front camera image. The system leverages a pre-trained semantic segmentation model \cite{chen2017rethinking}, that extracts semantic information from the input image, as pixel-wise labels in the scene, including roads, sky, and obstacle objects such as vehicles and road curbs. 
We transform obstacle locations to ground \RAL{plane} coordinates through a homography \cite{hartley2003multiple}, as visualized in the left plot in Figure~\ref{fig:utility}, \RAL{by assuming the obstacle can be approximated as a 2-D shape on the ground plane}. 
In parallel, we obtain a \emph{driver intention utility map} by computing a Gaussian kernel density estimation (KDE) \cite{parzen1962estimation} map extracted from a set of sampled predicted trajectories, as visualized in the middle plot in Figure~\ref{fig:utility}, \RALF{since the observed future trajectory is not available at test time. Additionally, the future trajectory ignores the multi-modality of human actions.} As an alternative, our driver intention utility map represents the marginalized estimate of the intended trajectory trace.

After obtaining the safety utility map and the driver intention utility map, we compute the ground truth utility using Eq.~\eqref{eq:utility_traj} given a trajectory sample.

\subsubsection{Backup Trajectory Planner}
\label{subsec:planner}
In order to guarantee driver safety in cases where the predicted driver behavior is risky, CARPAL leverages a planner to plan backup trajectories that are safe and follow driver intentions as close as possible when safety is guaranteed. 

The planner is designed to maximize the utility based on the utility maps from Section~\ref{sec:utility_computation}. Existing parallel autonomy approaches \cite{schwarting2017parallel,reddy2018shared} assume access to perfect environment information, which may fail in real systems. In contrast, to capture the inherent planner uncertainty due to the noise from upstream perception tasks, we add random obstacles or remove existing obstacles to and from the safety utility map, respectively, to simulate noises from perception and semantic segmentation systems. Besides, we shift the goal position, which can be obtained from training data, occasionally by a random noise vector, as the planner does not always have the perfect goal information from driver.

Given the safety utility map and the driver intention utility map, we combine them to generate a set of physically feasible shortest path to the goal state using a hybrid A* search algorithm \cite{dolgov2008practical} given vehicle specs, such as the starting velocity and turning radius. \RALF{We choose hybrid A* because of its simplicity, but it can be replaced by any other planners that reason about utility functions in a spatial map.} The output includes a set of planned backup trajectories given noise samples in perception and goal locations.


\subsubsection{Utility Approximator}
\label{subsec:utility_regressor}
In order to enable real-time decision making at test time, CARPAL embeds a utility approximator to regress the utility statistics. This is achieved by adding a few linear layers after the embedding layer in the prediction network to generate regressed scalar utility values.
\RAL{We want to emphasize that regressor has no access to planned or predicted trajectories when regressing the utility values, making it suitable for real-time decision making.}


\begin{figure}
\end{figure}

\begin{algorithm}[t!]
\DontPrintSemicolon
 \SetKwInOut{Input}{input}\SetKwInOut{Output}{output}
 \Input{$X =$ ($c$: front camera image, $x$: vehicle states), $\tau_a$: observed future trajectory,\quad\quad\quad\quad\quad\quad\quad
 $\mathcal{M}$: current prediction model.}
 \Output{$\mathcal{M}'$: updated prediction model.}
   \tcc{Forward prediction}
Obtain the predicted trajectory distribution and the estimated utility statistics using the current model $\mathcal{M}$:
 $\hat{\mathcal{T}}, \hat{\mu}_H, \hat{\sigma}^2_H, \hat{\mu}_P, \hat{\sigma}^2_P = \mathcal{M}(X).$ \;
 Generate $n$ predicted trajectory samples $\mathcal{T}_H  \sim \hat{\mathcal{T}}$.\;
  \tcc{Ground truth utility computation}
 Compute a set of $m$ safety utility maps $\mathbf{M}_o$ given camera image $c$ with $m$ independent noises.\;
 Compute driver intention cost map $M_i = \textit{KDE}(\mathcal{T}_H)$ given predicted trajectory samples.\;
 Compute the final cost maps $\mathbf{M} = \mathbf{M}_o + \alpha M_i$.\;
 Generate $m$ planned backup trajectories $\mathcal{T}_P$ given $\mathbf{M}$.\;
 Compute ground truth utility statistics $\mu_H, \sigma^2_H, \mu_P, \sigma^2_P$ given $\mathcal{T}_H, \mathcal{T}_P$.\;
   \tcc{Loss computation \& optimization}
 Compute losses: 
 $\mathcal{L}_{\textit{nll}} = -\log(P(\tau_a|\hat{\mathcal{T}})),$
 $\mathcal{L}_{\mu_H} = (\mu_H - \hat{\mu}_H)^2, \quad
 \mathcal{L}_{\sigma^2_H} = (\sigma^2_H - \hat{\sigma}^2_H)^2, $
 $\mathcal{L}_{\mu_P} = (\mu_P - \hat{\mu}_P)^2, \quad
 \mathcal{L}_{\sigma^2_P} = (\sigma^2_P - \hat{\sigma}^2_P)^2.$\;
 Update model to $\mathcal{M}'$ given losses and an optimizer. \;
 \caption{Training procedure of CARPAL.}
 \label{alg:training}
\end{algorithm}

\subsection{Model Losses}
We train our confidence-aware prediction model with five loss functions: a negative log-likelihood loss to measure the accuracy of predictions compared to the observed future trajectory, and four L2 losses between the regressed utility values and the target ground truth values. 
\subsection{Training and Test}
The training procedure of CARPAL is illustrated in Algorithm~\ref{alg:training}. At test time, we use the best trained model to predict future driver trajectories, as well as utility statistics for both predicted and planned trajectories, to enable real-time decision making in the context of parallel autonomy.

\subsection{Confidence-Aware Decision Maker}
While there exist many options for decision making depending on safety requirements and driver preferences, we propose a task-specific binary decision function $F$ in Algorithm~\ref{alg:decision} to support safe driving while avoiding false alarms.

\begin{figure}
\end{figure}
\begin{algorithm}[t!]
\DontPrintSemicolon
 \SetKwInOut{Input}{input}\SetKwInOut{Output}{output}
 \Input{$\mu_H, \sigma^2_H, \mu_P, \sigma^2_P, \eta_H, \eta_P$.}
 \Output{$d$: a binary decision bit, where $0$ means a negative decision (e.g., not intervene), and $1$ means a positive decision (e.g., intervene).}
        \uIf{($\mu_H < \mu_P$)}{
            \uIf{($\sigma^2_H < \eta_H$ \text{and} $\sigma^2_P < \eta_P$)}{
            \Return{1}
            }
            \Else{
            warn; \Return{0}
            }
        }
        \Else{
        \Return{0}
        }
 \caption{An example CARPAL decision function $F$.}
 \label{alg:decision}
\end{algorithm}

To guarantee driver safety, $F$ first compares the expected utility between the driver predictions ($\mu_H$) and a backup planner ($\mu_P$). In cases where $\mu_H$ is worse compared to $\mu_P$ (e.g., the predicted human trajectories are getting too close to obstacles), $F$ decides whether to intervene or not based on the uncertainty levels of the utility estimates as variances. If both utility uncertainties are small such that they are below certain thresholds $\eta_H$ and $\eta_P$ for human predictions and backup planner trajectories, respectively, $F$ outputs a positive decision (1) to intervene because of a high confidence level. For the sake of simplicity, we assume that $\eta_H = \eta_P$ and unify the notations as $\eta$.
Otherwise, $F$ simply warns to avoid false alarms. 
We use this extra logic to prevent interventions where there is a considerable chance of the driver utility outperforming the planner utility, and express people's willingness to accept human errors but not automated system ones.
We will show examples that merit such warning logic in Section~\ref{sec:qualitative}, yet it will not play a role in our quantitative analysis for our binary decision making system in Section~\ref{sec:quantitative}.
We defer reasoning about a ternary decision function as future work.

In contrast, if the expected driver utility is acceptable, $F$ outputs a negative decision 0 as to not intervene. 

Instead of comparing the expected driver utility $\mu_H$ to a fixed threshold to detect near collisions, we find it effective to compare it with the expected planner utility $\mu_P$, thanks to the design of our utility function in Eq.~\eqref{eq:utility_all} that includes a safety term and an driver intention term. In cases where $\mu_P$ is higher than $\mu_H$, the planner trajectories need to be farther away from the obstacles than the predicted trajectories to achieve a higher safety utility, as the intention utility $u_H$ would always favor the predicted driver trajectories. Furthermore, since the obstacle distance is transformed by a sigmoid function in the utility computation, the gap in $d^2_O(\cdot)$ would only make a difference if the distance from the driver trajectories to the obstacles is small enough. Otherwise, this distance would be in the saturation area of the sigmoid function, and thus lead to a similar utility compared to the utility of planner trajectories with a larger distance. 

\subsection{Bounding Utility Uncertainty in Parallel Autonomy}
Here we show that under certain conditions, the utility uncertainty of the parallel autonomy system is bounded by the utility uncertainties estimated by CARPAL. We denote the overall parallel autonomy utility entropy to be $h(U)$.The overall utility $U$ is a choice between the utility from human driver action $H$ and the utility from the planner action $P$.

\begin{lem}
For the purpose of deciding on intervention, and assuming the utility regressors estimate utility uncertainty with a margin of $\Delta_u$, the utility uncertainty of the parallel autonomy system is approximated by the regressor.
\end{lem}
\begin{proof}
Using chain rule for differential entropy:
\begin{equation}
    h(U) = h(U|H,P) + h(H,P) - h(H,P|U).
\end{equation}
Since $U$ can be completely determined by $H$ and $P$, we have $h(U|H,P) = 0$. In addition, since $U$ is a function of $H$ and $P$, we have $h(H,P|U)$ to be non-negative. Therefore,
\begin{equation}
\label{eq:entropy_1}
    h(U) = h(H,P) - h(H,P|U) \leq h(H,P).
\end{equation}
By properties of joint entropy, we get:
\begin{equation}
\label{eq:entropy_2}
    h(U) \leq h(H,P) \leq h(H) + h(P).
\end{equation}
\RAL{
We obtain the final result based on Gaussian approximations:
\begin{align}
    h(H) &= \ln (\sqrt{2\pi \sigma^2_H}) + \frac{1}{2} +\Delta_u,\\
    h(P) &= \ln (\sqrt{2\pi \sigma^2_P}) + \frac{1}{2} +\Delta_u.
\end{align}}
\end{proof}

\section{Results}
\label{sec:results}
In this section, we introduce the details of our method and two baseline methods in the context of a decision making problem for parallel autonomy, followed by a description of an augmented test set including risky scenarios used for verifying the intervention performance. We then demonstrate the advantage of our model over the baselines by measuring recall and fall-out rates in a binary decision setting, and show a number of intervention and warning use cases to highlight the effectiveness of our system.

To evaluate CARPAL and baselines, we introduce the following binary decision making problem $\mathcal{P}$ for parallel autonomy: Given future driver trajectory predictions and relevant statistics, a decision making system decides whether to intervene the driver. The decision can either result in a \emph{positive event}, in which intervention should happen when the observed future driver trajectory would reach a distance to the obstacles smaller than $d_S = 1.6$ meters (based on the size of a normal size car) and when the ground truth planner utility is better than the ground truth driver utility, or a \emph{negative event}, in which the intervention should not happen.
While most existing literature \cite{schwarting2017parallel,anderson2012constraint,erlien2015shared} define risky scenarios based on near collisions, our work also considers the reliability of the backup planner, which can sometimes fail due to perception errors, before an intervention should happen.
This allows us to ensure that the intervention is effective, as we expect the backup planner to improve the utility in risky scenarios.

\subsection{Model Details}
Our model utilizes a network similar to \cite{huang2019uncertainty}. We add a utility approximator that consists of three linear layers with (64, 16, 4) neurons, where each linear layer is followed by a batch norm, ReLU, and dropout layers. We trained our model using a Tesla V100 GPU in PyTorch.
The prediction horizon is 3 seconds. The weighting coefficient $\alpha$ in Eq.~\eqref{eq:utility_all} is 0.1. \RALF{The utility map is discretized at approximately 0.02 meters, to balance between planning time and performance.} The number of predicted samples $n$ and planning samples $m$ are 10.
\RAL{We trained the model for 2000 epochs with a batch size of 64 and a learning rate of 0.0001, and pretrained a prediction model for 1000 epochs without the utility estimation part.}

\RAL{The model was trained and validated on a naturalistic driving dataset collected by a test vehicle equipped with cameras, Global Positions System (GPS), and Inertial Measurement Unit (IMU). \RALF{The data consists of 0.6 million samples from 14 trips with a total duration of 30 hours in multiple urban trips under different weather and lighting (e.g. rainy, shady) conditions.} More details can be found at \cite{huang2019uncertainty}.}

\subsection{Task-Agnostic Baselines}
We introduce two baselines for solving $\mathcal{P}$. The first baseline assumes a constant velocity model, as commonly seen in shared control literature \cite{alonso2014shared}, to estimate risk. The second baseline is extended from \cite{huang2019uncertainty} and decides according to regressed task-agnostic accuracy of predicted driver trajectories. 

\subsubsection{Velocity-Based Predictor} 
The velocity-based predictor (VBP) assumes a constant velocity model to predict a deterministic future trajectory of the driver vehicle given the current velocity. To determine near collisions, VBP checks if the minimum distance (see Eq.~\eqref{eq:utility_obs}) between its prediction and any obstacles identified in the front camera image is smaller than a minimum safe distance threshold $d_S$. The system intervenes if a near collision is detected, and does nothing otherwise. While we could have used other dynamics models such as constant acceleration \cite{schwarting2017parallel}, we observe that a constant velocity model leads to better prediction accuracy over longer horizons and choose it as a representative from the existing parallel autonomy literature. 

\subsubsection{Accuracy-Based Predictor} 
The accuracy-based predictor (ABP) generates decisions by estimating the accuracy of the predicted trajectories, as done in \cite{huang2019uncertainty}. Accuracy is defined by the displacement error between the predicted trajectories and the observed driver trajectory. We compare the estimated prediction error with a threshold $\eta_{ABP}$, and if the predictor is accurate enough, we decide whether to intervene using the same criteria in VBP based on the minimum distance to the obstacles.
ABP is task-agnostic as it ignores the detailed task information and uses only the accuracy to decide on intervention choices. This is less effective in improving safety performance, as we will show in both quantitative and qualitative results.

\subsection{Augmented Risky Test Set}
We use a set of test examples to verify the intervention performance in the context of parallel autonomy.
As driving in this dataset was conducted by a safety driver, it does not include any risky scenarios, making it challenging to evaluate the performance of a system when intervention is needed.
Therefore, we augment 10\% of the test set using one of the two following ways to create \RAL{realistic ``risky'' counterfactual scenarios (not seen in training data)}.
First, we scale up the past and future trajectory of the driver by 20\%, causing the future trajectory to occasionally hit obstacles. This simulates cases where the driver is inattentive to the road and driving recklessly. 
Second, we add random obstacles to the front camera \RAL{RGB image input with a color similar to building walls}, to simulate risky events when a pedestrian or a cyclist jumps in front of our car.
We note these approaches work well because of the inputs of the predictor and planner we have chosen, \RAL{serving as reproducible surrogates for risky events data. In general, creating or collecting realistic risky events remains a difficult problem.}

\subsection{Quantitative Results}
\label{sec:quantitative}
\begin{figure}[t!]
    \centering
    \includegraphics[width=1.0\columnwidth]{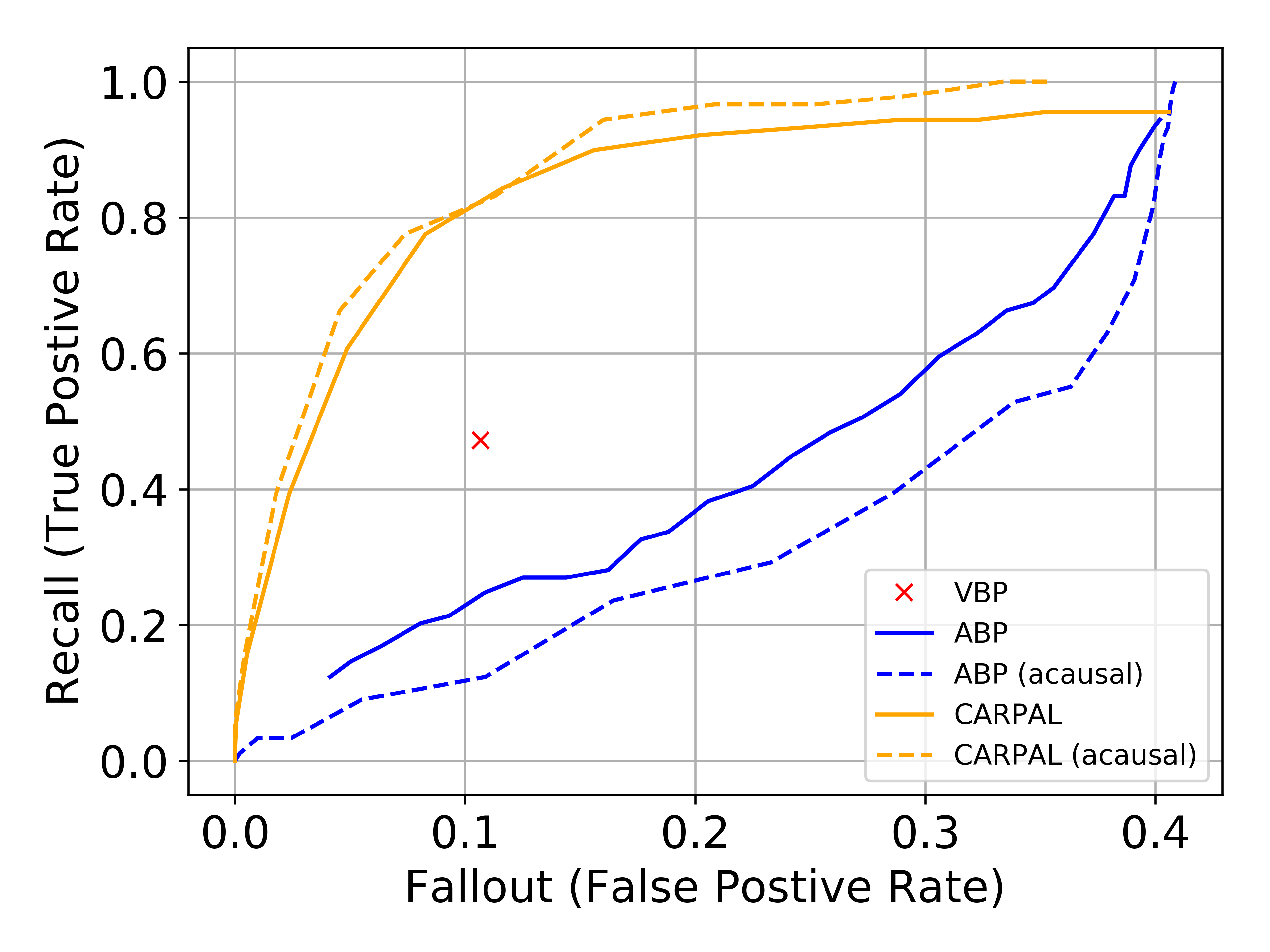}
    \caption{ROC curves for our method (CARPAL) and two baselines, where our method stably achieves high recall and low fall-out, as compared to the baselines. Solid lines indicate results using regressed utility estimates, while dashed lines indicate results using acausal utilities computed from data.}
    \label{fig:roc}
\end{figure}

To quantitatively evaluate CARPAL and baselines, we define two metrics: recall and fall-out, to measure the accuracy of our system and the false alarm rate, respectively. The first metric, recall (or sensitivity), measures the percentage of positive events that are recognized as positive and intervened by the system. The second metric, fall-out, measures the percentage of negative cases that are intervened. Our goal is to achieve high recall and low fall-out. 

Figure~\ref{fig:roc} shows the receiver operating characteristic (ROC) curves in solid lines for CARPAL and ABP by ranging possible threshold values, such as $\eta$ for our method and $\eta_{ABP}$ for ABP, and a fixed point, visualized as a red cross marker, for VBP since it does not depend on any thresholds. In addition to the two solid ROC curves, we plot two dashed ROC curves representing ``acausal'' CARPAL and ABP, using ground truth acausal utility values computed from offline data, to compare to the performance of our approximator and decision making approach that leverage only causal data.

\begin{figure*}[t!]
\begin{minipage}{0.33\textwidth}
\begin{subfigure}[b]{0.8\columnwidth}
    \includegraphics[width=5.8cm]{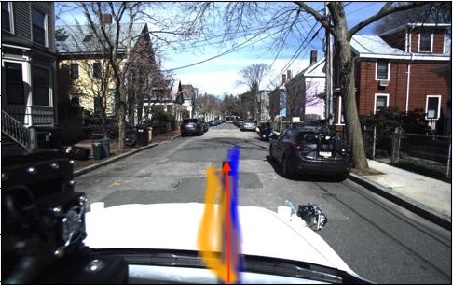}
  \end{subfigure}
  \vspace{1mm}
\\
\begin{subfigure}[b]{0.8\columnwidth}
    \includegraphics[width=5.8cm]{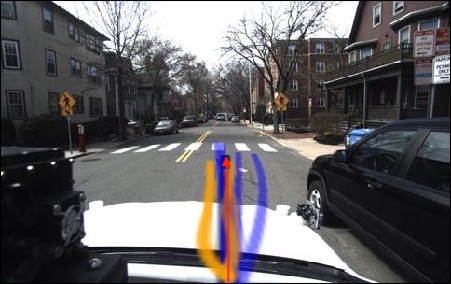}
  \end{subfigure}
  \begin{center}
  (a) No intervention.
  \end{center}
\end{minipage}
\begin{minipage}{0.33\textwidth}
\begin{subfigure}[b]{0.8\columnwidth}
    \includegraphics[width=5.8cm]{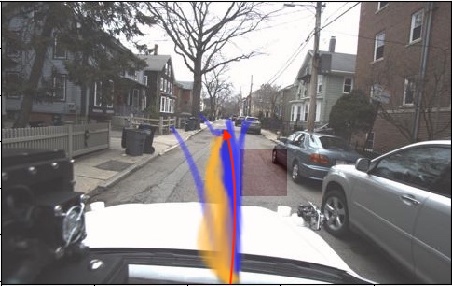}
  \end{subfigure}
  \vspace{1mm}
\\
\begin{subfigure}[b]{0.8\columnwidth}
    \includegraphics[width=5.8cm]{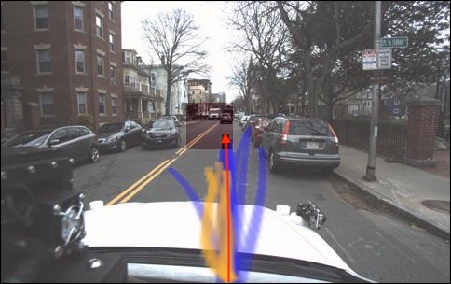}
  \end{subfigure}
  \begin{center}
  (b) Intervention.
  \end{center}
\end{minipage}
\begin{minipage}{0.33\textwidth}
\begin{subfigure}[b]{0.8\columnwidth}
    \includegraphics[width=5.8cm]{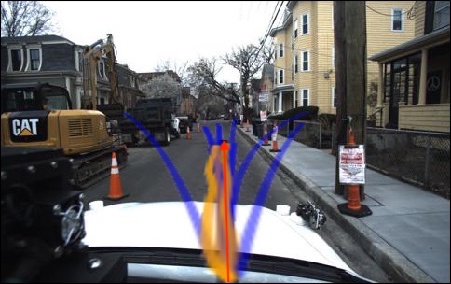}
  \end{subfigure}
  \vspace{1mm}
\\
\begin{subfigure}[b]{0.8\columnwidth}
    \includegraphics[width=5.8cm]{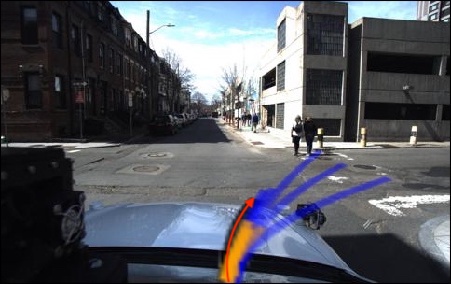}
  \end{subfigure}
  \begin{center}
  (c) Warning.
  \end{center}
\end{minipage}
  \caption{Visualizations of predictions (blue) and planned trajectories (orange) of different decisions made by CARPAL. 
  \RALF{Red arrow} indicates the observed future trajectory. Shaded squares indicate augmented obstacles. (a) Safe scenarios, where estimated driver utilities are high with small uncertainties, and CARPAL does not intervene. (b) Intervention scenarios, where the observed trajectories are risky. CARPAL intervenes after estimating the driver utilities to be low and the planner utilities to be high, both with high confidence levels. In contrast, the task-agnostic ABP baseline acts conservatively without intervention because the estimated prediction L2 distance errors are large. (c) Warning scenarios, where estimated driver utilities are low with high uncertainties. Even though the planner has better utilities, CARPAL warns instead of intervening to avoid false alarms.}
  \label{fig:qualitative_examples}
\end{figure*}



\textbf{Discussion:}
While both our method and ABP, shown in solid lines, trade off recall and fall-out at different thresholds, our method demonstrates better performance. In many positive cases where intervention is needed, ABP becomes conservative and fails to intervene due to low estimated prediction accuracy. On the other hand, our method is able to utilize the utility estimates, especially by recognizing cases where the planner utility is higher than the driver utility, to correctly identify such positive cases. Additionally, we see that a deterministic predictor such as VBP would outperform ABP due to ABP's conservatism. However, VBP suffers from its prediction accuracy and has higher fallout compared to our method at the same recall level. This is because VBP, relying on the current velocity, does not have enough information to recognize that drivers may slow down when approaching obstacles in the near future, and thus creates more false alarms. \RALF{On average, our system improves the utility by 35.6\% in risky scenarios.}

When comparing between solid lines and dashed lines, we observe that although our model estimates the utility values based on only causal data, its performance is close to an acausal model with access to ground truth utilities computed using information from offline data. The imperfect acausal ROC curve for CARPAL is due to the inaccuracies from the driver intention utility map -- ideally, it should should sit at the top left corner with 0 fallout and 1 recall if perfect driver intention is given.
On the other hand, the acausal ABP performs slightly worse than ABP, due to the fact that task-agnostic measures fail to consider their impact on downstream tasks, which lead to worse performance even though the ground truth utility values are provided.

In addition to its online performance, CARPAL takes approximately 9.25 milliseconds to regress utility estimates for each data sample \RAL{on a Tesla V100 GPU (and 40 milliseconds on a laptop Quadro P2000 GPU)}, making it applicable in real-time decision making systems. On the other hand, it takes approximately half a second to compute actual utility statistics for 10 trajectories following procedures in Section~\ref{sec:utility_computation}. The comparison further validates the advantage of our utility approximator in terms of time complexity.



\subsection{Qualitative Results}
\label{sec:qualitative}

We present a handful of examples in Figure~\ref{fig:qualitative_examples} to demonstrate three possible use cases of CARPAL in the context of parallel autonomy. In safe scenarios, we want to avoid overtaking the driver to maintain the usability of our system. In other scenarios where predicted trajectories have low utilities, we should gauge the estimated utility uncertainties to decide whether to intervene or warn, in order to achieve a high recall rate while avoiding false alarms.

\subsubsection{Safe Scenarios}
In most daily driving scenarios, the driver behaves safely and intervention is not needed. We show two common safe cases in the left column of Figure~\ref{fig:qualitative_examples}, where the drivers operate the vehicle safely on an open road or at an open intersection. Our system estimates high utilities on the driver predictions with low uncertainties, and thus decides to do nothing, regardless of the planner utilities. 

\subsubsection{Intervention Scenarios}
In cases where the driver actions are risky, our system is able to recognize them and intervene after estimating a low driver utility and a high planner utility with high confidence. In addition to the motivating example in Figure~\ref{fig:motivating_example}, we illustrate two extra examples shown in the middle column of Figure~\ref{fig:qualitative_examples}, in which both driver actions, indicated by the observed future trajectories in red, lead to a collision with an augmented obstacle. The obstacle in the top example simulates a situation where a pedestrian jumps suddenly behind a parked car, and the obstacle in the bottom example simulates a vehicle or a cyclist on the opposite lane trying to pass the car blocking the road.
In both cases, our system decides to intervene after, with high estimated confidence levels, estimating a low utility for the predicted driver trajectories in blue and a high utility for the planner trajectories in yellow that can be used to take over driver control and avoid collisions.

The illustrated examples also demonstrate the advantage of our system over a task-agnostic system such as ABP, which decides whether to intervene based on the estimates over prediction accuracies and tends to be conservative by not intervening in these cases since the estimated accuracies are low. Instead of relying on a fixed threshold against prediction accuracies, our system approximates the utilities in the downstream backup planning task and finds them to be better than the average predicted driver utilities with high confidence levels, leading to better decisions.

\subsubsection{Warning Scenarios}
In addition to intervening the drivers when a clear near collision is detected, we demonstrate a few examples where intervention is unneeded, but a warning can be helpful to remind the driver of potential risks.
In the examples shown in the right column of Figure~\ref{fig:qualitative_examples}, our system detects that the driver utilities are lower than the planner utilities. However, due to a novel input with construction cones in the top example or the existence of moving pedestrians, the estimated driver utilities come with high uncertainties. Therefore, our system decides to warn the driver instead of taking over control. This would allow our system to successfully prevent a false alarm, as indicated by the safe observed trajectory, while notifying the driver potential risks. 
\section{Conclusions}
\label{sec:conclusions}
In this paper, we propose a multi-task trajectory predictor that estimates in real-time the utility statistics associated with its predictions for downstream tasks, and show how our utility estimates benefit a parallel autonomy system. Compared to task-agnostic methods, our online system achieves a higher true positive rate while maintaining a lower false alarm rate in an augmented test set, considering common risky driving intervention use cases. 
\RALF{Future work includes considering different utility and decision functions, validation in a closed-loop simulation benchmark, and demonstration in a live footage.}




\bibliographystyle{IEEEtran}
\bibliography{ref}

\end{document}